\documentclass[a4paper,conference]{IEEEtran}
\pdfoutput=1
\IEEEoverridecommandlockouts

\usepackage{cite}
\usepackage{amsmath,amssymb,amsfonts}
\usepackage{algorithmic}
\usepackage{graphicx}
\usepackage{textcomp}

\usepackage{xspace}
\usepackage[dvipsnames]{xcolor}
\usepackage{booktabs}
\usepackage{hyperref}
\newcommand{\ours}{DIAG\xspace}

\makeatletter
\DeclareRobustCommand\onedot{\futurelet\@let@token\@onedot}
\def\@onedot{\ifx\@let@token.\else.\null\fi\xspace}
 
\def\ie{i.e\onedot}

\makeatother

\def\BibTeX{{\rm B\kern-.05em{\sc i\kern-.025em b}\kern-.08em
    T\kern-.1667em\lower.7ex\hbox{E}\kern-.125emX}}

\makeatletter
\def\ps@IEEEtitlepagestyle{%
  \def\@oddfoot{\mycopyrightnotice}%
}
\def\mycopyrightnotice{%
\begin{minipage}{\textwidth}
\centering \footnotesize
Copyright~\copyright~2024 IEEE. Personal use of this material is permitted.  Permission from IEEE must be obtained for all other uses, in any current or future media, including reprinting/republishing this material for advertising or promotional purposes, creating new collective works, for resale or redistribution to servers or lists, or reuse of any copyrighted component of this work in other works.
\end{minipage}
}
\makeatother
\begin{document}

\title{Leveraging Latent Diffusion Models for Training-Free In-Distribution Data Augmentation for Surface Defect Detection
\thanks{
This study was carried out within the PNRR research activities of the consortium iNEST (Interconnected North-Est Innovation Ecosystem) funded by the European Union Next-GenerationEU (Piano Nazionale di Ripresa e Resilienza (PNRR) – Missione 4 Componente 2, Investimento 1.5 – D.D. 1058  23/06/2022, ECS\_00000043).
This manuscript reflects only the Authors’ views and opinions.
Neither the European Union nor the European Commission can be considered responsible for them.
Furthermore, this study was also partially funded by the European Union - NextGenerationEU (Italiadomani, Piano Nazionale di Ripresa e Resilienza (PNRR) - Missione 4 - Componente 2, Investimento 3.3 - progetto M4C2).
}
}

\author{
    \IEEEauthorblockN{Federico Girella, Ziyue Liu, Franco Fummi, Francesco Setti, Marco Cristani, Luigi Capogrosso}
    \IEEEauthorblockA{\textit{Department of Engineering for Innovation Medicine, University of Verona, Italy}}
    {\tt name.surname@univr.it}
}

\maketitle

\begin{abstract}
Defect detection is the task of identifying defects in production samples.
Usually, defect detection classifiers are trained on ground-truth data formed by normal samples (negative data) and samples with defects (positive data), where the latter are consistently fewer than normal samples.
State-of-the-art data augmentation procedures add synthetic defect data by superimposing artifacts to normal samples to mitigate problems related to unbalanced training data.
These techniques often produce out-of-distribution images, resulting in systems that learn what is not a normal sample but cannot accurately identify what a defect looks like.
In this work, we introduce \ours{}, a training-free Diffusion-based In-distribution Anomaly Generation pipeline for data augmentation.
Unlike conventional image generation techniques, we implement a human-in-the-loop pipeline, where domain experts provide multimodal guidance to the model through text descriptions and region localization of the possible anomalies.
This strategic shift enhances the interpretability of results and fosters a more robust human feedback loop, facilitating iterative improvements of the generated outputs. 
Remarkably, our approach operates in a zero-shot manner, avoiding time-consuming fine-tuning procedures while achieving superior performance. 
We demonstrate the efficacy and versatility of \ours{} with respect to state-of-the-art data augmentation approaches on the challenging KSDD2 dataset, with an improvement in AP of approximately 18\% when positive samples are available and 28\% when they are missing.
The source code is available at \url{https://github.com/intelligolabs/DIAG}.
\end{abstract}


\section{Introduction}
\label{sec:intro}

\begin{figure}
\centering
\includegraphics[width=\linewidth]{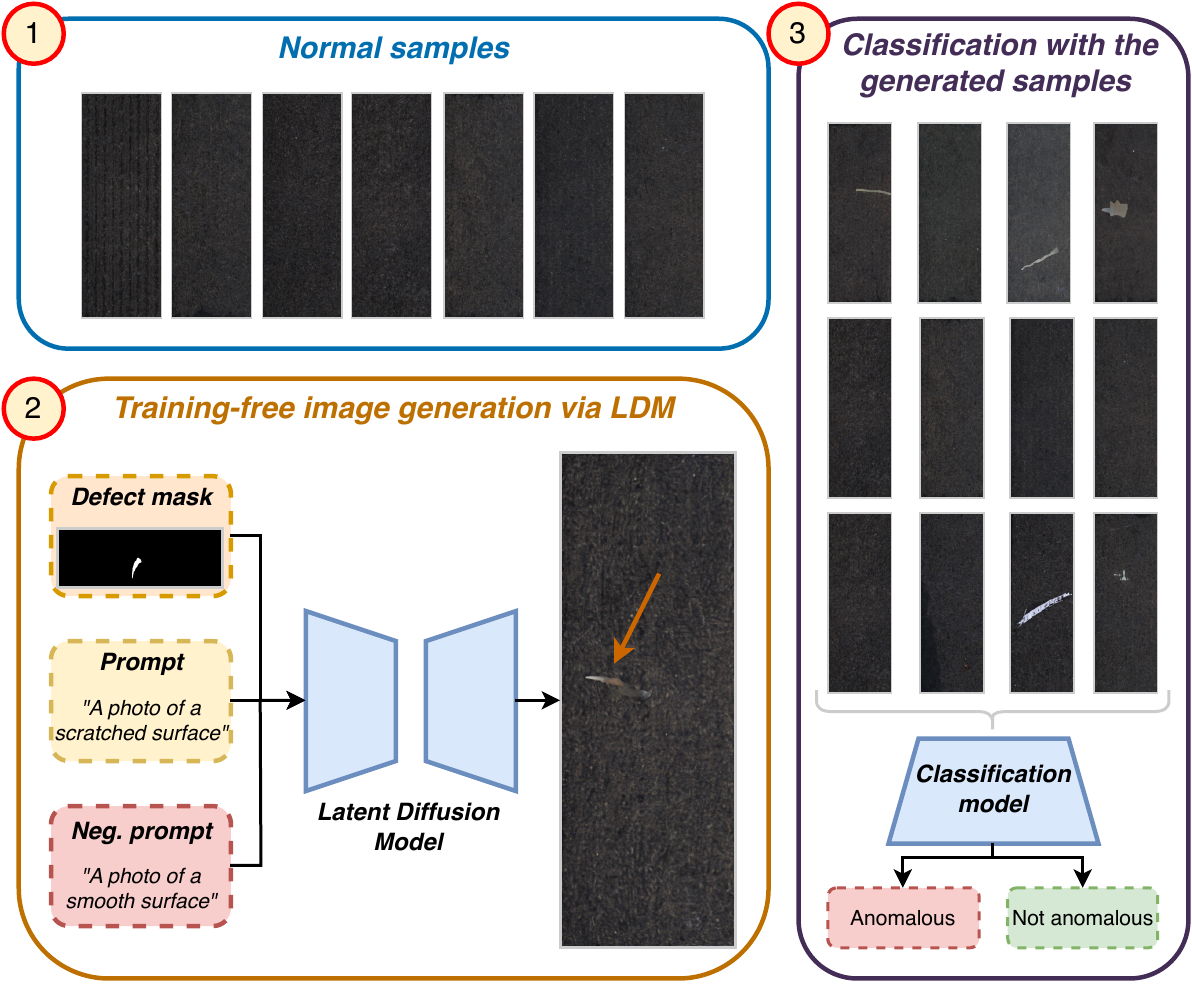}
\caption{The \ours{} pipeline.
Starting from positive samples, we leverage a Latent Diffusion Model (LDM) to synthesize novel in-distribution high-quality images of defective surfaces based on defect localization and textual prompts.
These synthetic images are then used as anomaly samples to train a binary classifier for anomaly detection.}
\label{fig:teaser}
\end{figure}

Surface Defect Detection (SDD) is a challenging problem in industrial scenarios, defined as the task of individuating samples containing a defect~\cite{wang2018fast}, \ie{}, samples that do not conform to a prototypical texture.
In many real-world applications, a human expert inspects every product and removes those defective pieces.
Unfortunately, training human experts can be expensive. 
Moreover, humans are relatively slow in accomplishing this task, and their performances are subject to stress and fatigue.

Automated defect detection systems~\cite{tsang2016fabric,hanzaei2017automatic} can easily overcome most of these issues by learning classifiers on defective and nominal training products.
The main drawback is the data collection process required to train a model effectively.
Indeed, defective items (\ie{}, positive samples) are relatively rare compared to nominal items (\ie{}, negative samples).
Thus, the user may need to collect massive amounts of data to have enough positive samples.
Moreover, with the rise of the Industry 4.0 paradigm and the transition towards flexible manufacturing processes, there is an increasing demand for systems that can quickly adapt to new production setups~
\cite{capogrosso2024machine}, \ie{}, customized products manufactured in small batches.
Traditional automated systems cannot comply with these demands since data collection could easily involve the whole batch size.

Recent studies on SDD focused on limiting the impact of the labeling process by formulating the problem under the unsupervised learning paradigm~\cite{deng2022anomaly,roth2022towards} or training exclusively on nominal samples~\cite{rudolph2021same}, possibly using few-shot learning strategies~\cite{song2023comprehensive}.
In both cases, the goal is to generate an accurate model of the nominal sample distribution and classify everything with a low probability score as anomalies.
However, due to the limited restoration capability of these models, these approaches tend to generate many false positives, especially on datasets with complex structures or textures~\cite{chen2021surface}.

It is worth noting that, in industrial setups, anomalies are not generated by Gaussian processes but are the outcome of specific, often predictable, issues during the production process.
Consequently, the anomalous samples are not randomly distributed outside the nominal distribution; they can be modeled as a mixture of Gaussian distributions in the feature space instead.
Expert operators can easily define the main problems they can expect from the manufacturing process, such as which kind of defects, in which locations, and how often they expect them to appear.
Thus, generative AI can represent a powerful tool for SDD, with defect image generation emerging as a promising approach to enhance detector performance.
Specifically, in recent years, Denoising Diffusion Probabilistic Models (DDPMs)~\cite{ho2020denoising} received significant attention as a powerful class of generative models.

In this paper, we propose an interactive learning protocol where a vision language model is used to generate realistic images starting from textual prompts.
Specifically, we promote using DDPMs to produce fine-grained realistic defect images that can be used as positive samples to train an anomaly detection model.
We name our approach \ours{}, a training-free \underline{D}iffusion-based \underline{I}n-distribution \underline{A}nomaly \underline{G}eneration pipeline for data augmentation in the SDD task.
By leveraging pre-trained DDPMs with multimodal conditioning, we can exploit domain experts' knowledge to generate plausible anomalies without needing real positive data.
When using these augmented images to train an anomaly detection model, we show a notable increase in the detection performance compared to previous state-of-the-art augmentation pipelines.
Figure~\ref{fig:teaser} outlines our approach, which will be explained in Section~\ref{sec:method}.

The main contributions of our work are as follows:
\begin{itemize}
\item We present \ours{}, a complete pipeline for training anomaly detection models based on nominal images and textual prompts.
We showcase the superior outcomes achieved by utilizing generated defective samples compared to previous state-of-the-art approaches.
\item We dive into spatial control approaches to enable the synthesis of defect samples incorporating regional information and exhibit enhanced controllability of the image generation through a human-in-the-loop pipeline, effectively utilizing domain expertise to generate more plausible in-distribution anomalies.
\end{itemize}

\section{Related Work}
\label{sec:related}

Surface defect detection refers to identifying and categorizing irregularities, flaws, or imperfections on the surface of materials or objects.
These defects include scratches, cracks, discolorations, and any other anomaly that deviates from the expected surface quality.
Research has been conducted according to different setups: unsupervised approaches~\cite{tao2022unsupervised} use a mixture of unlabelled positive and negative sample images for training; supervised approaches require labeled samples in the form of binary masks representing the defects (full supervision)~\cite{luan2020siamese} or simply as a tag for the whole image (weak supervision)~\cite{bovzivc2021mixed}.
Supervised methods demonstrated superior accuracy in the identification of anomalies.
Nevertheless, the effort required to provide good annotations is not always justified.
Collecting positive samples can be time and resource-consuming due to the low rate of defective products generated by industrial lines.

Thus, many recent approaches adopt a ``\emph{clean}'' setup, where the training set consists of only nominal samples.
Two strategies can be adopted in clean setups: model fitting and image generation.
Model fitting approaches aim at generating an accurate model of the nominal distribution, considering an outlier every sample with a likelihood lower than --or a distance from the nominal prototype higher than-- a predefined threshold~\cite{rudolph2021same,defard2021padim}.
On the contrary, data augmentation approaches leverage generative methods to synthesize images of defects and use these images as positive samples for training a supervised model.
Specifically, this work focuses on generation-based data augmentation under clean setups.

The most popular data augmentation pipeline for SDD consists of a series of random standard transformations of the input image --such as mirroring, rotations, and color changes-- followed by the super-imposition of noisy patches~\cite{yang2023memseg}.

In~\cite{zavrtanik2021draem}, an ablation study focused on the generation of synthetic anomalies leads to the following findings: \textit{i)} adding synthetic noise images is never counterproductive, it just diminishes the effectiveness in percentage; \textit{ii)} few generated anomaly images (in the order of tens) are enough to increase the performance substantially; \textit{iii)} textural injection in the anomalies is essential, or, equivalently, adding uniformly colored patches is ineffective.

In MemSeg~\cite{yang2023memseg}, the pipeline for the generation of the abnormal synthetic examples is divided into three steps: \textit{i)} a Region of Interest (ROI) indicating where the defect will be located is generated using Perlin noise and the target foreground; \textit{ii)} the ROI is applied to a noise image to generate a noise foreground ROI; \textit{iii)} the noise foreground ROI is super-imposed on the original image to obtain the simulated anomalous image.
However, all these approaches are based on generating out-of-distribution patterns that do not faithfully represent the target-domain anomalies. 

In~\cite{zhang2023prototypical}, the authors introduced the concept of ``extended anomalies'', where the specific abnormal regions of the seen anomalies are placed at any possible position within the normal sample after applying random spatial transformations. Unfortunately, this requires an accurate segmentation of the training images, an operation we may want to avoid. 

More recently, the first work that draws attention to in-distribution defect data is In\&Out~\cite{capogrosso2024diffusion}, in which the authors empirically show that diffusion models provide more realistic in-distribution defects. 

In this paper, we significantly improve the generation of in-distribution anomalous samples incorporating domain knowledge provided by an expert user through textual prompts and localization of salient regions.
We will use state-of-the-art MemSeg~\cite{yang2023memseg} and In\&Out~\cite{capogrosso2024diffusion} methods as competitors for our augmentation pipeline.
With \ours{}, we can produce a distribution of defective images closer to the real one, resulting in a more precise identification of the decision boundaries in the classification step.

\section{Methodology}
\label{sec:method}

In this section, we provide detailed explanations of \ours{}.
In particular, Section~\ref{method:background} covers techniques for diffusion-based image generation, Section~\ref{subsec:image_gen} showcases the anomalous image generation pipeline, and Section~\ref{subsec:anomaly_training} outlines the anomaly detection model training procedure.

\subsection{Multimodal diffusion-based image generation} \label{method:background}
DDPMs~\cite{sohl2015deep,ho2020denoising} are a class of deep latent variable models that work by modeling the joint distribution of the data over a Markovian inference process.
This process consists of small perturbations of the data with a variance-preserving property~\cite{song2020score}, such that the limit distribution after the diffusion process is approximately identical to a known prior distribution.
Starting with samples from the prior, a reverse diffusion process is learned by gradually denoising the sample to resemble the initial data by the end of the procedure.

Formally, the data distribution $q(x_0)$ is modelled through a latent variable model $p_\theta(x_0)$:
\begin{gather}
p_\theta(x_0) = \int p_\theta(x_{0:T}) dx_{1:T}\;,\\ \quad  p_\theta(x_{0:T}) := p_\theta(x_T) \prod_{t=1}^{T} p^{(t)}_\theta(x_{t-1} | x_t)\;,
\label{eq:gen}
\end{gather}
where $x_1, \ldots, x_T$ are latent variables of the same dimensionality as $x_0$.

The parameters $\theta$ are learned by maximizing an ELBO of the log evidence, \ie{}:
\begin{gather}
\max_\theta \mathbb{E}_{q(x_0)}[\log p_\theta(x_0)] \leq \nonumber \\
\max_\theta \mathbb{E}_{q(x_0, x_1, \ldots, x_T)}\left[\log p_\theta(x_{0:T}) - \log q(x_{1:T} | x_0) \right]\;,
\label{eq:elbo}
\end{gather}
where $q(x_{1:T} | x_0)$ represents a fixed inference process defined as the following as a Markov chain:
\begin{gather}
q(x_{1:T} | x_0) := \prod_{t=1}^{T} q(x_t | x_{t-1})\;,\\
q(x_t | x_{t-1}) := \mathcal{N}\left(\sqrt{\frac{\alpha_t}{\alpha_{t-1}}} x_{t-1}, \left(1 - \frac{\alpha_t}{\alpha_{t-1}}\right) I\right)\;,
\label{eq:diff-ho}
\end{gather}
where $\alpha_{1:T} \in (0, 1]^T$ is a predefined variance schedule, and the covariance matrix is ensured to have positive terms on its diagonal. 
Specifically, this parametrization has the property:
\begin{align} \label{eq:ddpm-xt-parametrization}
q(x_t | x_0) = \int q(x_{1:t} | x_0) dx_{1:(t-1)} = \nonumber \\
\mathcal{N}(x_t; \sqrt{\alpha_t} x_0, (1 - \alpha_t)I)\;,
\end{align}
therefore we can write $x_t$ as a linear combination of $x_0$ and a noise variable $\epsilon$. 

When we set $\alpha_{T}$ sufficiently close to $0$, $q(x_T | x_0)$ converges to a standard Gaussian for all $x_0$, so it is natural to set $p_\theta(x_T) := \mathcal{N}(0, \mathbf{I})$.
Given that all the conditionals are modeled as Gaussians with fixed variance, the objective in Equation~\eqref{eq:elbo} can be greatly simplified. 
In particular,~\cite{ho2020denoising} shows that the following (further simplified) lower bound provides optimal generative performance:
\begin{gather} \label{eq:ddpm-loss}
L(\epsilon_\theta) := \sum_{t=1}^{T} \mathbb{E}_{x_0,\epsilon_t}\left[ \lVert {\epsilon_{\theta}^{(t)}(\sqrt{\alpha_t} x_0 + \sqrt{1 - \alpha_t} \epsilon_t) - \epsilon_t}\rVert_2^2 \right]\;,
\end{gather}
where $x_0\sim q(x_0), \epsilon_t \sim \mathcal{N}(0, I)$, $\epsilon_\theta = \{\epsilon_\theta^{(t)}\}_{t=1}^{T}$ is a set of $T$ functions, with each $\epsilon_{\theta}^{(t)}: X \to X$ having trainable parameters $\theta^{(t)}$.

In practice, these functions are approximated by a neural network conditioned on the diffusion time $t$. 
After the model is trained, we can generate new samples by first sampling $x_T$ from the known prior $p_\theta(x_T)$, and then iteratively reversing the diffusion process, thereby sampling $\{x_{T-1}\ldots x_0 \}$.

In addition, we leveraged the natural ability of DDPMs to incorporate multimodal conditioning in the generation process, taking inspiration from~\cite{ho2022classifier,rombach2022high,zhang2023adding,capogrosso2023neuro}.
Specifically, we will use prompts, \ie{}, textual descriptions of the anomaly, and negative prompts, i.e.,  prompts that guide the image generation ``away'' from its concepts. 
This results in high-quality images that comply with the given descriptions~\cite{saharia2022photorealistic,ramesh2022hierarchical,podell2023sdxl}.

In particular, we opt to utilize an inpainting model, as demonstrated in~\cite{sohl2015deep,rombach2022high}.
Given an image with a masked region, inpainting seamlessly fills it with content that harmonizes with the surrounding image.
Although typically employed to eliminate undesired artifacts, the inpainting process ensures that the masked area incorporates the provided prompt, effectively merging textual and visual content.

\subsection{The \ours{} pipeline} \label{subsec:image_gen}
To generate an anomalous image $i_a$, the process starts by sampling a random negative image, an anomaly description, and a mask, forming the triplet $(i_n, d_a, m_a)$.
Instead of directly operating on the image pixels using DDPM, we use a Latent Diffusion Model (LDM) to work in a lower-dimensional latent space~\cite{rombach2022high}.
Thus, the above information will be fed to a text-conditioned LDM to perform inpainting on image $i_n$ using the mask $m_a$.

The anomaly description $d_a$ guides the generation, filling the masked region of $i_n$ with an anomaly that complies with the prompt.
To generate images resembling real anomalous samples, domain knowledge from industrial experts is exploited, providing textual descriptions of the potential anomalies' type, shape, and spatial information.

The LDM is then conditioned on this information to inpaint plausible anomalies on defect-free samples.
Formally, given pictures of defect-free (negative) samples $I_n$, domain experts will provide textual descriptions $D_a$ of what different anomalies may look like.
At the same time, regions where these anomalies may appear on the defect-free samples will be designated.
We define this set of regions as a set of binary masks $M_a$ of possible anomalies, shapes, and locations.
The result of this operation is $i_a$, an anomalous version of $i_n$, where an anomaly has been inpainted in the masked region $m_a$.
Due to the stochastic nature of LDMs, this process can be repeated multiple times to generate an augmented set of anomalous sample images $I_a$.
Finally, the set $I_a$ can be used as data augmentation for training anomaly detection models, as presented in the following section.

\subsection{Anomaly detection task} \label{subsec:anomaly_training}
We approach the anomaly detection problem as a binary classification problem, where the objective is to predict whether a sample belongs to one of two classes.
Specifically, we utilized a ResNet-50~\cite{he2016deep} backbone trained with a binary cross-entropy loss function denoted as $\mathcal{L}_{\text{BCE}}$.
The binary cross-entropy loss measures the dissimilarity between the predicted probability distribution and the actual distribution of the labels.
Mathematically, it is defined as:
\begin{equation}
\mathcal{L}_{\text{BCE}}(y, \hat{y}) = -\frac{1}{N} \sum_{i=1}^{N} \left[ y_i \log(\hat{y}_i) + (1 - y_i) \log(1 - \hat{y}_i) \right]\;,
\end{equation}
where, $y$ represents the ground truth labels, $\hat{y}$ represents the predicted probabilities, and $N$ is the number of samples.
In detail, $y_i$ denotes the true label for sample $i$, which can be either 0 or 1, while $\hat{y}_i$ signifies the predicted probability that sample $i$ belongs to class 1.

\section{Experiments}
\label{sec:experiments}

In this section, we show the efficacy of our data augmentation approach for defect detection from a quantitative and qualitative point of view.

\subsection{Experiment setup}

\paragraph*{\textbf{Datasets}}
We use the Kolektor Surface-Defect Dataset 2 (KSDD2)~\cite{bovzivc2021mixed}, one of the most recent, complex, and real-world surface defect detection datasets.
This dataset comprises 246 positive and 2085 negative images in the training set and 110 positive and 894 negative images in the testing set.
Positive images are images with visible defects, such as scratches, spots, and surface imperfections.

Since the images have different dimensions, we standardize the dataset resolution, resizing all the images to $224\times{}632$ pixels while keeping the number of normal and anomalous samples unchanged.

\paragraph*{\textbf{Evaluation metrics}}
The anomaly detection performance was evaluated based on Average Precision (AP), Precision, and Recall, following the evaluation protocol defined in~\cite{capogrosso2024diffusion}.

Additionally, to evaluate the visual similarity between generated images and the original dataset images, we employ the Fréchet Inception Distance (FID)~\cite{heusel2017gans}, a popular metric in the image generation field which computes the distance between the distribution of two sets of images.
More specifically, the Fréchet distance calculates distance $d(.,.)$ between a Gaussian with mean $(m, C)$ obtained from $p(.)$ and a Gaussian with mean $(m_w, C_w)$ obtained by $p_w(.)$, where $p_w(.)$ represents real world data and $p(.)$ represents generated data.
In practice, these distributions are two sets of data: the ``world'' data (\ie{}, the images in a dataset) and the ``generated'' data (\ie{}, the generated images).
These sets are then fed to an Inception model pre-trained on ImageNet to extract deep features from each sample of the distributions.
The resulting two sets of features represent the Gaussians with mean $(m_w, C_w)$ and $(m, C)$ for the ``world'' and ``generated'' data, respectively.
Specifically,~\cite{heusel2017gans} shows that a lower FID score matches a human's higher perceived visual similarity (a lower perceptual distance), meaning that similar sets of images will have a lower FID than dissimilar sets.
Formally, the FID score is:
\begin{equation}
d((m, C), (m_w, C_w))=||m-m_w||^2_2 +\\ \text{T}(C + C_w - 2(CC_w)^\frac{1}{2})
\end{equation}
where $\text{T}$ refers to the trace linear algebra operation.

\subsection{Implementation details}

In this section, we specify all the implementation details for reproducibility.
All training and inferences were conducted on an NVIDIA RTX 3090 GPU.

\paragraph*{\textbf{Inpainting via \ours{}}}
We use the pre-trained implementation of SDXL~\cite{podell2023sdxl} from Diffusers~\cite{diffusers} as our text-conditioned LDM.
In particular, SDXL shows drastically improved performance compared to the previous versions of Stable Diffusion~\cite{rombach2022high} and achieves results comparable to commercial state-of-the-art image generators.

Following the procedure outlined in Section~\ref{subsec:image_gen}, we use the negative images of KSDD2 as the set $I_n$.
As the set of anomaly descriptions $D_a$, we used the prompts ``\texttt{white marks on the wall}'' and ``\texttt{copper metal scratches}''. 
Instead, ``\texttt{smooth, plain, black, dark, shadow}'' were used as a negative prompt to improve the performance further.
These prompts were selected through a human-in-the-loop iterative pipeline, until the resulting images resembled plausible anomalies.
We used the segmentation masks of positive samples in the KSDD2 dataset to simulate the domain experts' definition of plausible anomalous regions.

Then, these data are fed to the pre-trained SDXL model to perform inpainting on the negative images in a training-free process, generating the set of augmented anomalous images $I_a$ as described in Section~\ref{subsec:image_gen}.

Finally, the generated images $I_a$ are added to the training set, which will be used to train the anomaly detection model.

\paragraph*{\textbf{ResNet-50 training and testing}}
For a fair comparison with~\cite{capogrosso2024diffusion}, we use the same PyTorch implementation of the ResNet-50~\cite{he2016deep} as our anomaly detection model, in which we substitute the fully connected layers after the backbone to make it a binary classifier.
The network is trained for 50 epochs with Adam~\cite{kingma2014adam} as an optimizer, a learning rate of $0.0001$, and a batch size of 32.

To maintain consistency with the training and evaluation procedures of KSDD2, our setup is the same as presented in~\cite{bovzivc2021mixed,capogrosso2024diffusion}, where only the images and ground truth labels are used to train the model.
For our comparison, we use the official code of In\&Out~\cite{capogrosso2024diffusion} to fine-tune a DDPM in the \textit{full-shot} scenario (on all the positive images of the KSDD2 dataset) and generate their augmented images.
Likewise, we follow the procedure of MemSeg~\cite{yang2023memseg} to generate the ``per-region'' augmented images. 
Finally, we generate \ours{} augmented images, following the inpainting methodology outlined in Section~\ref{sec:method}.
The set of images used for training changes depending on the experiment and the pipelines being tested, but in general, it can be seen as a combination of the original negative images $I_n$, an optional set $I_p$ of original positive images, and the set of generated positive images $I_a$.

\subsection{Quantitative results} \label{subsec:quantitative_results}

\paragraph*{\textbf{Zero-shot data augmentation}} \label{par:zero_shot}
\begin{table}[t!]
\centering
\caption{Results between MemSeg, In\&Out and \ours{} when \textbf{\textit{no}} anomalous samples are available.
In \textbf{bold}, the best results. \underline{Underlined}, the second best.}

\begin{tabular}{l|c|c|c|c}
\toprule
\textbf{Model} & $\mathbf{N_{aug}}$ & \textbf{AP $\uparrow$} & \textbf{Precision $\uparrow$} & \textbf{Recall $\uparrow$} \\
\midrule
MemSeg~\cite{yang2023memseg}  & 80   & .514         & .733          & .436 \\
MemSeg~\cite{yang2023memseg}  & 100  & .388         & .633          & .432 \\
MemSeg~\cite{yang2023memseg}  & 120  & .511         & .683          & .470 \\
\midrule
In\&Out~\cite{capogrosso2024diffusion}       & 80  & .556          & .530          & .655 \\
In\&Out~\cite{capogrosso2024diffusion}       & 100 & .626          & .742          & .568 \\
In\&Out~\cite{capogrosso2024diffusion}       & 120 & .536          & .699          & .534 \\
\midrule
\ours{} (ours)                          & 80  & \underline{.769} & .851             & \textbf{.673} \\
\ours{} (ours)                          & 100 & \textbf{.801}    & \underline{.924} & \underline{.664} \\
\ours{} (ours)                         & 120 & .739             & \textbf{.944}    & .609 \\
\bottomrule
\end{tabular}
\label{tab:zero_shot}
\end{table}

Here, we emulate the situation where \textbf{\textit{no}} original positive samples are available in the training set.
This scenario makes generating augmented positive samples necessary and restricts the users to augmentation procedures that do not rely on positive images.
To do this, we build the set of augmented anomalous images $I_a$ by generating $N_{aug}$ augmented positive samples with different pipelines, \ie{}, MemSeg~\cite{yang2023memseg}, In\&Out~\cite{capogrosso2024diffusion} and \ours{}.
Then, we train a ResNet-50 on a dataset that includes the original negative samples $I_n$ and the augmented positive samples $I_a$.
Finally, we evaluate the model on the original test set.

Table~\ref{tab:zero_shot} reports the comparison between the models trained with MemSeg, In\&Out, and \ours{} augmented data at different values of $N_{aug}$.
As we can see, our proposed method achieves the highest AP (.801), recorded at 100 augmented images, while also resulting in a consistently higher AP  when compared to the MemSeg and In\&Out pipelines.
These impressive results highlight how, through domain expertise in the form of anomaly descriptions and segmentation masks, it is possible to generate in-distribution images able to meaningfully guide an anomaly detection network, even in a complicated scenario where no real anomalous data is available.

Surprisingly, the \ours{} performance with $N_{aug}=120$ augmented images is lower than using a smaller number of augmented images.
We hypothesize this is due to the stochastic nature of the LDMs image generation.
While it allows the generation of various images given the same guidance, it can also lower, in some cases, the predictability of the quality of the generated samples, which sometimes may not faithfully comply with the prompt. 
Future works will focus on studying quality consistency in the image generation pipeline.

\paragraph*{\textbf{Full-shot data augmentation}}
\begin{table}[t!]
\centering
\caption{Results between MemSeg, In\&Out and \ours{} when \textbf{\textit{all}} the anomalous samples are available.
In \textbf{bold}, the best results. \underline{Underlined}, the second best.}

\begin{tabular}{l|c|c|c|c}
\toprule
\textbf{Model} & $\mathbf{N_{aug}}$ & \textbf{AP $\uparrow$} & \textbf{Precision $\uparrow$} & \textbf{Recall $\uparrow$} \\
\midrule
MemSeg~\cite{yang2023memseg}  & 80   & .744 & .851 & .691 \\
MemSeg~\cite{yang2023memseg}  & 100  & .774 & .814 & .752 \\
MemSeg~\cite{yang2023memseg}  & 120  & .734 & .772 & .707 \\
\midrule
In\&Out~\cite{capogrosso2024diffusion}       & 80  & .747 & .764 & .734 \\
In\&Out~\cite{capogrosso2024diffusion}       & 100 & .775 & .868 & .720 \\
In\&Out~\cite{capogrosso2024diffusion}       & 120 & .782 & .906 & .689 \\
\midrule
\ours{} (ours)                       & 80     & .869             & \underline{.912} & .755 \\
\ours{} (ours)                          & 100    & \underline{.911} & \textbf{.978}    & \underline{.800} \\
\ours{} (ours)                          & 120    & \textbf{.924}    & .896             & \textbf{.864} \\
\bottomrule
\end{tabular}
\label{tab:full_shot}
\end{table}

To showcase \ours{} as a general data augmentation technique, we also explore the scenario where real positive samples are available in the training set.
To this aim, we include all the 246 real positive samples $I_p$ in the training set, together with the real negative images $I_n$ and the $N_{aug}$ augmented positive images $I_a$.

As we can see from Table~\ref{tab:full_shot}, \ours{} achieves the highest average AP yet (.924), surpassing the .782 set by the previous state-of-the-art data augmentation pipeline~\cite{capogrosso2024diffusion}.
When comparing these results to the ones obtained in the ``zero-shot data augmentation'' scenario, it is clear how more in-distribution images improve model performance during training.
This is highlighted by the improvement in performance of all the models when adding the real positive images $I_p$ to the training set.
At the same time, the inclusion of \ours{} augmented images allows the model to explore the anomaly distribution further, resulting in the difference in performance between the different data augmentation pipelines.

\subsection{Qualitative results} \label{subsec:qualitative_results}
\begin{figure*}[t!]
\centering
\includegraphics[width=\linewidth]{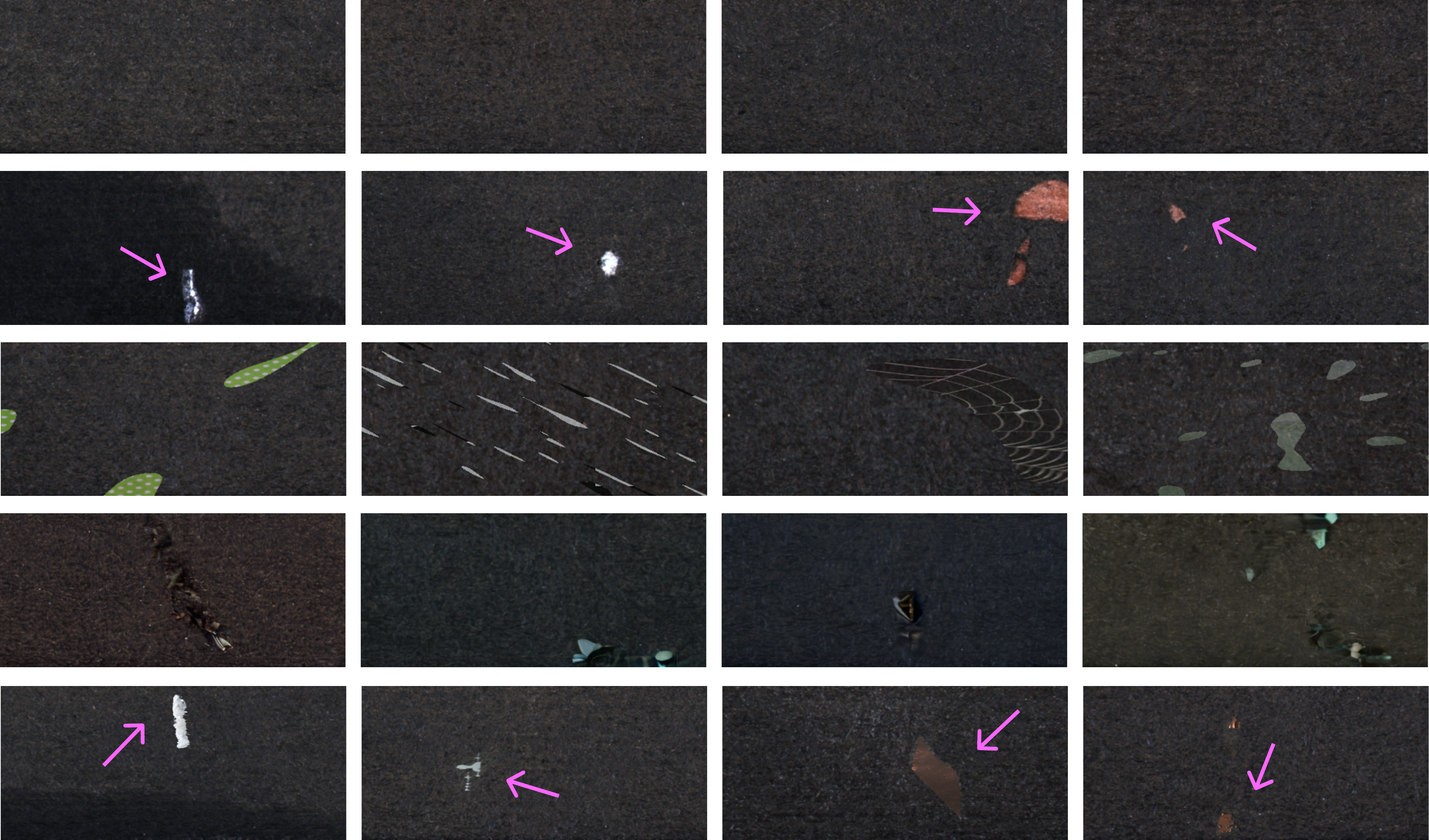}
\caption{First row displays some negative samples from the KSDD2 dataset.
Instead, the second row shows some images of positive samples from the same dataset.
In the third row, we show the MemSeg-generated defect samples.
The fourth row shows In\&Out generated defect samples.
Lastly, the final row showcases images generated with \ours{}.
Notably, the defect images that \ours{} generated are more realistic and in-distribution.}
\label{fig:generated_images}
\end{figure*}

The main goal of our data augmentation pipeline is to generate in-distribution synthetic positive images, meaning images that closely resemble the real ones.
Figure~\ref{fig:generated_images} shows qualitative results.
It's evident that the images produced by \ours{} are markedly more realistic compared to those generated by MemSeg~\cite{yang2023memseg} and In\&Out~\cite{capogrosso2024diffusion}.

In addition, we provide a numeric evaluation of the similarity between the generated images and the real ones by employing FID~\cite{heusel2017gans}.
It is worth noting that due to the limited number of anomalous images in the original dataset, we are forced to calculate FID on a different network layer, precisely the second max pooling layer.
This is a common procedure in cases where the number of images is low, as the calculation requires the number of samples (images) to be higher than the number of features.
Note that this only changes the magnitude of the values obtained, not the metric's general behavior.
In the specific case of KSDD2, we choose the first and second max-pooling layers with 64 and 192 features, respectively.
Specifically, we compare the images generated with MemSeg, In\&Out, and \ours{} with the ones available in the KSDD2 dataset and compute the FID scores between the positive images of the KSDD2 and the previously mentioned sets of augmented images.

\begin{table}[t!]
\centering
\caption{FID scores between the real positive images of KSDD2 and the images generated by MemSeg, In\&Out and \ours{}. 
The scores are calculated using the first and second max pooling layers of the Inception network, having 64 and 192 features, respectively.
In \textbf{bold}, the best results.}

\begin{tabular}{l|c|c}
\toprule
\textbf{Augmentation procedure} & \textbf{FID 64 $\downarrow$}  & \textbf{FID 192 $\downarrow$}\\
\midrule
MemSeg~\cite{yang2023memseg}  & 0.834   & 4.376 \\
DDPM~\cite{capogrosso2024diffusion} & 0.334 & 1.520 \\
\ours{} (ours)   & \textbf{0.096} & \textbf{0.411} \\
\bottomrule
\end{tabular}
\label{tab:fid}
\end{table}
The results, reported in Table~\ref{tab:fid}, highlight how \ours{} can generate images that are very similar to the ones originally present in the dataset, resulting in the lowest FID out of all the other methodologies.

Another interesting observation is how both the generative-model-based procedures (DDPM and \ours{}) result in images that are more in-distribution (lower FID) than the ``per-region'' augmentation techniques such as MemSeg, which records the highest FID out of all the tested methodologies.

\section{Conclusions}
\label{sec:conclusions}

This work presents \ours{}, a novel data augmentation pipeline that leverages language-conditioned Latent Diffusion Models to produce training-free positive images for enhancing the performance of a surface defect detection model.
We introduced domain experts in the generation pipeline, asking them to describe with textual prompts how a defect should look and where it can be localized.
Then, we adopt a pre-trained LDM to generate defective images and train a binary classifier for isolating the anomalous images.
We focus our experiments on the KSDD2 dataset and establish ourselves as the new state-of-the-art data augmentation pipeline, surpassing previous approaches in both the zero-shot and full-shot data augmentation scenarios with an AP of .801 and .924, respectively.
These results highlight the potential of in-distribution data augmentation in the anomaly detection field, where training-free generative model pipelines such as \ours{} can provide meaningful data for downstream classification, making them appealing solutions in scenarios where real anomalous data is difficult to collect or unavailable.
These promising results promote further exploration across various datasets, particularly investigating how robust the image generation is compared to noisy textual prompts.

\bibliographystyle{IEEEtran}
\bibliography{bibi}

\end{document}